\pdfoutput=1

\documentclass[11pt]{article}

\usepackage[]{acl}

\usepackage{times}
\usepackage{latexsym}

\usepackage[T1]{fontenc}

\usepackage[utf8]{inputenc}

\usepackage{microtype}

\usepackage{microtype}

\usepackage{subfigure}
\usepackage{graphicx}
\usepackage{lipsum}
\usepackage{microtype}
\usepackage{float}
\usepackage{algorithm}
\usepackage{algorithmic}
\usepackage{placeins}
\usepackage{booktabs}
\usepackage{cleveref}
\usepackage{multirow}
\usepackage{comment}
\usepackage{amssymb}
\usepackage{stackengine}
\usepackage{scalerel}
\usepackage{xcolor}

\newcommand\openbigstar[1][0.7]{%
  \scalerel*{%
    \stackinset{c}{-.125pt}{c}{}{\scalebox{#1}{\color{white}{$\bigstar$}}}{%
      $\bigstar$}%
  }{\bigstar}
}

\title{An Isotropy Analysis in the Multilingual BERT Embedding Space}

\author{Sara Rajaee$^{\openbigstar[0]}$ \and Mohammad Taher Pilehvar$^{\openbigstar[.5]}$\\
  $~^{\openbigstar[0]}$Iran University of Science and Technology, Tehran, Iran \\
  $~^{\openbigstar[.5]}$Tehran Institute for Advanced Studies, Khatam University, Iran \\
  \texttt{sara{\_}rajaee@comp.iust.ac.ir}\\
  \texttt{mp792@cam.ac.uk}}

\begin{document}
\maketitle
\begin{abstract}
Several studies have explored various advantages of multilingual pre-trained models (such as multilingual BERT) in capturing shared linguistic knowledge. However, less attention has been paid to their limitations.
In this paper, we investigate the multilingual BERT for two known issues of the monolingual models: anisotropic embedding space and outlier dimensions.
We show that, unlike its monolingual counterpart, the multilingual BERT model exhibits no outlier dimension in its representations while it has a highly anisotropic space. There are a few dimensions in the monolingual BERT with high contributions to the anisotropic distribution. However, we observe no such dimensions in the multilingual BERT.
Furthermore, our experimental results demonstrate that increasing the isotropy of multilingual space can significantly improve its representation power and performance, similarly to what had been observed for monolingual CWRs on semantic similarity tasks.
Our analysis indicates that, despite having different degenerated directions, the embedding spaces in various languages tend to be partially similar with respect to their structures.\footnote{Our code and datasets are publicly available at: \url{https://github.com/Sara-Rajaee/Multilingual-Isotropy}}.

\end{abstract}

\section{Introduction}

The multilingual BERT model \citep[mBERT]{devlin-etal-2019-bert}, pre-trained on 104 languages with no supervision, has shown impressive ability in capturing linguistic knowledge across different languages ~\citep{pires-etal-2019-multilingual}. 
Many studies have explored the encoded knowledge in multilingual CWRs using probing tasks and under zero-shot setting ~\citep{wu-dredze-2019-beto,K2020Cross-Lingual,chi-etal-2020-finding}.  
Following the probing studies, in this paper, we investigate the multilingual embedding space of BERT, focusing on its geometry in terms of isotropy. Previous research has shown that many pre-trained models, such as GPT-2 ~\citep{radford2019language}, BERT, and RoBERTa ~\citep{Liu2019RoBERTaAR}, have degenerated embedding spaces that downgrade their semantic expressiveness ~\citep{ethayarajh-2019-contextual,cai2021isotropy,rajaee-pilehvar-2021-cluster}. Several proposals have been put forward to overcome this challenge ~\citep{gao2018representation,zhang-etal-2020-revisiting}. 
However, to our knowledge, no study has so far been conducted on the degeneration problem in multilingual embedding spaces.
\begin{figure}[t!]

    \centering
    \includegraphics[width=3.55cm,height=2.75cm]{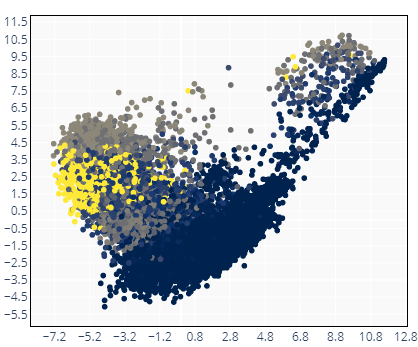}
    \includegraphics[width=3.55cm,height=2.75cm]{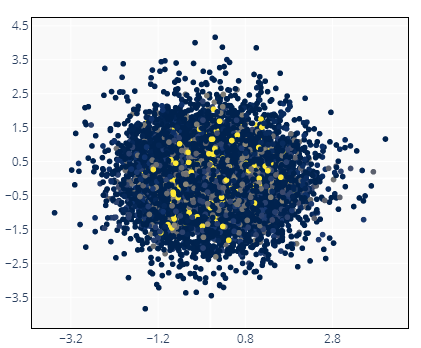}
    \caption{Degenerated (left) and isotropic (right) embedding spaces for Spanish plotted using PCA. Frequency-based distribution can be easily detected in the space (lighter colors indicate higher frequency). See Appendix \ref{appendix-freq} for more languages.}
    \label{fig:CWRs-distribution}

\end{figure}

Using two well-known metrics, we evaluate isotropy in the mBERT embedding space for six different languages (including two low resources): English, Spanish, Arabic, Turkish, Sundanese, and Swahili. 
We find that the representation spaces are massively anisotropic in all these languages.
However, unlike monolingual CWRs, where a few dimensions dominate the cosine similarity metric and have a high contribution to the anisotropic distribution \citep{timkey-van-schijndel-2021-bark}, there is no dominant dimension that defines anisotropy in the multilingual representations. 
Extending our study to other structural properties of the multilingual space, we also investigate outliers, i.e., specific dimensions with consistently high values \citep{kovaleva-etal-2021-bert}. 
Our findings reveal that, as opposed to English BERT, the multilingual BERT space does not involve any major outliers.
This indicates that the suggestion of \citet{luo-etal-2021-positional} on the role of positional embeddings in the emergence of outliers may not be valid.
Moreover, our analysis reveals that word frequency plays an important role in the distribution of the multilingual embedding space: words with similar frequencies create distinct local regions in the embedding space.

In analyzing multilingual space, we take a further step toward making the space isotropic. By applying a cluster-based isotropy enhancement method ~\citep{rajaee-pilehvar-2021-cluster}, we demonstrate that increasing isotropy of multilingual embedding space can result in significant performance improvements on semantic textual similarity tasks.
Our frequency analysis and the remarkable performance improvement in the zero-shot setting denote that the feature space of mBERT has encoded a common linguistic knowledge into its dominant directions to some extent across different languages.

\section{Background}
The representation degeneration problem in LMs has attracted lots of attention in recent years. Several regularizer-based methods have been proposed to make the space isotropic by adding an extra constraint to the loss function during pre-training ~\citep{gao2018representation,zhang-etal-2020-revisiting,Wang2020Improving}. Because of the re-training cost, other light approaches have been presented as a post-processing step \citep{li-etal-2020-sentence,rajaee-pilehvar-2021-cluster}. 
While analyzing the isotropy of embedding space is a well-studied area in English space, there are limited related studies on the multilingual embedding space. In this line, \citet{vulic-etal-2020-good} investigated the structural similarity of different language embedding spaces by evaluating their isomorphism. \citet{DBLP:journals/corr/abs-2107-09186} showed the positive effect of isotropic space on the degree of isomorphism, which in turn results in improved performance in cross-lingual alignment algorithms.
However, a focused study on the isotropy of multilingual embedding space has not been conducted. In this work, we provide more insights on the anisotropic distribution of multilingual embeddings and their notable differences from their English counterpart.

\subsection{Isotropy}


Geometrically, in an anisotropic space, embeddings occupy a narrow cone. 
This brings about an overestimation of the similarity between embeddings ~\citep{gao2018representation}. 
To quantify isotropy, we utilize two well-known metrics based on cosine similarity and principal components (PCs).

\paragraph{Cosine Similarity.}
\citet{ethayarajh-2019-contextual} used cosine similarity between random embeddings as an approximation of isotropy in the space. As mentioned before, random embeddings with an isotropic distribution have near-zero cosine similarities. The metric can be formulated as follows:

\begin{equation}
    \small
    I_{Cos}(\mathcal{W}) = \frac{1}{{N}} \sum_{i=1, x_i \neq y_i}^N Cos(x_i, y_i)
\end{equation}
where $x_i\in X, y_i\in Y$, $X$ and $Y$ are the sets of randomly sampled embeddings, and $\mathcal{W}$ is the embedding matrix. $N$ is the number of sampled pairs that is set to 1000 in our experiments. 
Lower $I_{Cos}(\mathcal{W})$ values indicate higher isotropy.

\paragraph{Principal Components.}
\citet{mu2018allbutthetop} proposed a metric based on principal components (PCs), approximated as follows:

\begin{equation}
    \small
    I_{PC}(\mathcal{W}) \approx \frac{\min_{u\in U}F(u)}{\max_{u\in U}F(u)}, F(u) = \sum_{i = 1}^{M} \exp(u^T w_i)\label{eq:1}
\end{equation}

\noindent where $w_i$ is the $i^{th}$ word embedding, $M$ is the number of all representations in the space, $U$ is the set of eigenvectors of the embedding matrix, and $F(u)$ is the partition function described in Equation \ref{eq:1}.  ~\citet{arora-etal-2016-latent} proved that $F(u)$ could be approximated using a constant for isotropic embedding spaces. Therefore, $I_{PC}(\mathcal{W})$ would be close to one in an isotropic embedding space.

\begin{table}
\centering
\setlength{\tabcolsep}{3.5pt}
\scalebox{0.72}{
\begin{tabular}{l c c c c c c c}
\toprule
\multicolumn{1}{c}{}            &
\multicolumn{1}{c}{\bf BERT} &
\multicolumn{6}{c}{\bf mBERT} \\

\cmidrule(lr){2-2}
\cmidrule(lr){3-8}

& {En} & {En} & {Es}  & {Ar} & {Tr} & {Su} & {Sw}\\ 
\midrule
$I_{Cos}(\mathcal{W})$ & 0.34  & 0.24 & 0.27 & 0.27 & 0.25 & 0.25 & 0.27     \\
$I_{PC}(\mathcal{W})$  &2.4E-5 & 6.4E-5 & 5.0E-5 & 1.6E-5 & 2.5E-4 & 1.2E-4 & 7.8E-5   \\ 

\bottomrule
\end{tabular}}
\caption{\label{isotropy-table} The isotropy of BERT and mBERT on a sub-set of Wikipedia, reporting based on $I_{Cos}(\mathcal{W})$ and $I_{PC}(\mathcal{W})$. }
\end{table}

\section{Analysis}

For all experiments, we opted for the multilingual BERT model (mBERT), which has a 12-layer transformer-based architecture similar to English BERT-base, and the representations are obtained from the last layer\footnote{To broaden our insights on the geometry of multilingual space, we expand our analyses to XLM-R model. The related results are reported in Appendix \ref{app:XML-R}.}. 
We selected English, Spanish, Arabic, Turkish, Sundanese, and Swahili. Our selection of these languages was to ensure that our analysis covers both high and low-resource languages.
As our evaluation benchmark, we chose a subset of Wikipedia articles in the selected languages. The analysis experiments in Sections \ref{sec:probing-isotropy} to \ref{sec:outliers} have been conducted on the same dataset.

In what follows, we first assess isotropy as a desirable property in the multilingual space and evaluate the contribution of individual dimensions to this property. 
We also expand our study to the outliers and word-frequency bias in CWRs. The former is a weak point of language models, and the latter is a well-known bias in the monolingual embedding space.
Lastly, we assess the effect of isotropy enhancement on the quality of the multilingual embeddings in the semantic similarity task.

\subsection{Probing isotropy}
\label{sec:probing-isotropy}
As the first step, we quantify the isotropy of the mBERT and BERT embedding spaces using the two metrics. 
For mBERT, we separately assess the isotropy of each language in the embedding space.

Based on the presented results in Table \ref{isotropy-table}, the average cosine similarity between random embeddings ($I_{Cos}(\mathcal{W})$) is much higher than zero, denoting anisotropic distribution in all considered languages. Measuring isotropy using $I_{PC}(\mathcal{W})$ also confirms the anisotropy issue in mBERT's space as well as the monolingual BERT model. 

Aligned with the numerical results, the illustration of multilingual CWRs in the left column of Figure \ref{fig:CWRs-distribution} gives us a clear perspective of the degenerated distribution in space.

\begin{figure*}[t!]
    \centering
    \includegraphics[width=16cm,height=6.1cm]{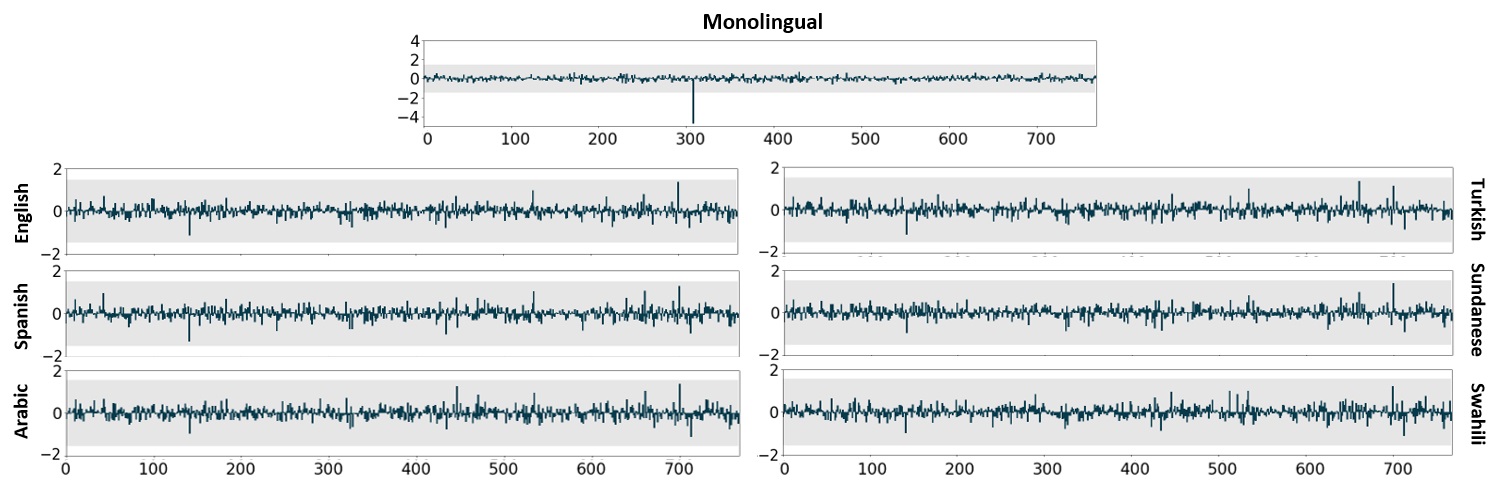}
    \caption{The average representation in English BERT (top) and mBERT (bottom). The shaded area denotes $3\sigma$. While an outlier has emerged in the former, we do not see any major outliers in the multilingual space.}
    \label{fig:Outliers}

\end{figure*}

\begin{table}[t!]
\centering
\setlength{\tabcolsep}{8pt}
\scalebox{0.8}{
\begin{tabular}{lc c ccc}
\toprule
\textbf{} & \textbf{$I_{Cos}(\mathcal{W})$} && \textbf{First} & \textbf{Second} & \textbf{Third} \\ 
\midrule
BERT           & 0.34       && 0.385      & 0.005      & 0.005     \\ 
\midrule
English        & 0.24       && 0.041      & 0.029      & 0.020      \\
Spanish        & 0.27       && 0.033      & 0.029      & 0.018     \\
Arabic         & 0.27       && 0.033      & 0.025      & 0.022      \\
Turkish        & 0.25       && 0.036      & 0.024      & 0.024      \\
Sundanese      & 0.25       && 0.036      & 0.016      & 0.016      \\
Swahili        & 0.27       && 0.025      & 0.018      & 0.014      \\

\bottomrule
\end{tabular}
}

\caption{\label{table-contr}The contribution of top-three dimensions to the expected cosine similarity ($I_{Cos}(\mathcal{W})$) in BERT and mBERT models.}
\end{table}

\subsection{Sensitivity to Rogue  Dimensions}
\label{sec:sensivity}
As we discussed before, cosine similarity is a widely used metric to measure the degree of isotropy in embedding space where a near-zero similarity demonstrates isotropic distribution. In this section, we evaluate the contribution of individual dimensions to the cosine similarity of two randomly chosen embeddings. 
Evaluating dimension-wise contribution sheds more light on the sensitivity of similarity-based metrics to individual dimensions and their role in (an)isotropic distribution.  

The dimension-based cosine similarity between $x$ and $y$ vectors with $d$ dimensions can be defined as follows \cite{timkey-van-schijndel-2021-bark}:
\begin{equation}
    Cos
    (x, y) = \sum_{i=1}^{d} \frac{x_i y_i}{\Vert x \Vert \Vert y \Vert}
\end{equation}

\noindent where $CC_i = x_i y_i/\Vert x \Vert \Vert y \Vert$ is the contribution of $i^{th}$ dimension to the cosine similarity.

We compute the average cosine similarity, $I_{Cos}(\mathcal{W})$, by randomly sampling 1000 token pairs and report the average contribution of the top-three dimensions to the average cosine similarity.

The results are reported in Table \ref{table-contr}. Aligned with the findings of \citet{timkey-van-schijndel-2021-bark}, we observe that only one dimension has a considerable contribution to the cosine similarity metric in the BERT embedding space. 
Therefore, the anisotropic distribution is dominated by the mentioned dimension and is not a global property of the whole space.
Unlike the monolingual BERT, multilingual BERT has no rogue dimensions. Hence, the anisotropic structure of the multilingual space cannot be attributed to certain dimensions.

\subsection{Outlier Dimensions}
\label{sec:outliers}
Pre-trained LMs exhibit consistent outliers, peculiar dimensions with large values, in their LayerNorms’ weights and consequently, in their contextual representations across all layers. Through several experiments, \citet{kovaleva-etal-2021-bert} have demonstrated that disabling these outliers can notably impair the performance of pre-trained and fine-tuned LMs.
Trying to find the root cause of outliers, \citet{luo-etal-2021-positional} have shown that removing positional embeddings makes the outliers disappear, concluding that the positional information is responsible for the emergence of outliers.

In this part, we investigate rogue dimensions in the multilingual embedding space of BERT. We check outliers by averaging over 10000 randomly selected representations and calculate the mean and standard deviation ($\sigma$) of dimensions’ distribution. 
Following the \citet{kovaleva-etal-2021-bert}’s suggestion, we consider a dimension as an outlier if its value is at least 3$\sigma$ larger/smaller than the mean of the distribution.

Results are shown in Figure \ref{fig:Outliers}.
On top, the outlier dimension of the mean representation can be easily detected in the English BERT. 
However, interestingly, multilingual BERT exhibits no major outliers in its embedding space across different languages. 
It can be concluded that, contrary to the suggestion of \citet{luo-etal-2021-positional}, positional embeddings cannot be responsible for outliers, given that both multi- and monolingual spaces are constructed using the same training procedure involving positional encodings.
We leave further investigation of outliers in contextual embedding spaces to future work.

Putting together the results of the previous sections, we observe that the mBERT embedding space is highly anisotropic, despite not having any outliers or dominant dimensions in the cosine similarity metric.

\subsection{Word frequency Bias}

\label{sec-frequency}
It has been shown that frequency plays an important role in the distribution of CWRs. 
Frequency-similar words make distinct local regions in the embedding space ~\citep{gao2018representation}, with high-frequency and rare words being around the center and far from the origin, respectively \cite{li-etal-2020-sentence}.
Frequency-based distribution is a factor that hampers the expressiveness of the embedding space. 
So, it is essential to investigate frequency bias in the multilingual embedding space.  

In this experiment, we analyze English, Spanish, and Arabic since there are enough resources to properly define word frequency in these languages. 
We randomly sample 500 sentences from the corresponding Wikipedia datasets and obtain a word representation by averaging over all its sub-token representations.

Figure \ref{fig:CWRs-distribution} shows the distribution of word representations per word frequency\footnote{We used the wordfreq library (\url{https://pypi.org/project/wordfreq/}). See Appendix \ref{app-wordfreq}.}. 
Every point represents a word embedding dyed based on its frequency.
As can be observed on the left, multilingual CWRs are biased toward their frequency, where words with similar frequencies create clustered regions. 
A similar pattern can be observed for the English BERT CWRs \citep{rajaee-pilehvar-2021-cluster}, with the only difference that in mBERT, low-frequency words are distributed near the origin and frequent words are far from it.

\begin{table*}[t]
\centering
\setlength{\tabcolsep}{7pt}
\scalebox{0.81}{
\begin{tabular}{lccccccc}
\toprule
  & \textbf{Ar-Ar} & \textbf{Ar-En} & \textbf{Es-Es} & \textbf{Es-En}  &\textbf{Es-En-WMT} & \textbf{Tr-En} & \textbf{En-En}  \\ 
\cmidrule(lr){2-2}
\cmidrule(lr){3-3}
\cmidrule(lr){4-4}
\cmidrule(lr){5-5}
\cmidrule(lr){6-6}
\cmidrule(lr){7-7}
\cmidrule(lr){8-8}
\bf Baseline     & 51.76 \textit{(8E-5)}  & 10.61 \textit{(1E-4)} 
                 & 64.15 \textit{(3E-5)}  & 31.26 \textit{(5E-4)} 
                 & 11.39 \textit{(1E-4)}  & 17.78 \textit{(1E-4)} 
                 & 60.82 \textit{(2E-6)}   \\
\midrule
\bf Individual   & 64.26 \textit{(0.60)}   &  23.10 \textit{(0.57)}  
                  & 70.88 \textit{(0.54)}  &  46.23 \textit{(0.50)}   
                  & 13.47 \textit{(0.50)}  &  25.59 \textit{(0.55)}
                  & 71.99 \textit{(0.54)}   \\

\bf Zero-shot     & 52.76 \textit{(6E-5)} & 19.36 \textit{(0.04)}
                  & 65.69 \textit{(8E-4)} & 43.82 \textit{(0.09)}
                  & 13.68 \textit{(8E-3)} & 19.89 \textit{(0.03)}       
                  &  -  \\
\bottomrule
\end{tabular}
}
\caption{\label{experiment-perofmance-tabel}STS performance (Spearman correlation percentage) on multi- and cross-lingual datasets using mBERT. Isotropy is reported based on $I_{PC}(\mathcal{W})$ in parentheses. Applying the cluster-based method can improve the performance on the multi- and cross-lingual datasets in both Individual and Zero-shot settings.}
\end{table*}
\subsection{Isotropy Enhancement}

Making the embedding space isotropic has theoretical and empirical benefits \citep{gao2018representation}. 
Several approaches have been proposed to improve isotropy in monolingual CWRs.
Some require a re-training of the model with additional objectives to address the degeneration problem \citep{gao2018representation, zhang-etal-2020-revisiting}, whereas others are applied as a light post-processing \cite{mu2018allbutthetop}.
To investigate the effect of isotropy enhancement for the multilingual embedding space, we opted for our cluster-based approach \cite{rajaee-pilehvar-2021-cluster}, which is a recent example from the latter category.
The proposed method splits the space into several clusters and discards dominant directions for each cluster.
The approach also allows us to investigate the similarity of the clustered structure of the embedding space across different languages under a zero-shot setting. 
More details on this method can be found in Appendix \ref{app-CIE}.
 
We consider the multilingual and cross-lingual Semantic Textual Similarity ~\citep[STS]{cer-etal-2017-semeval} that involves instances from Arabic, English, Spanish, and Turkish (Appendix \ref{task}). 
We run our experiments in \textbf{Individual} and \textbf{Zero-shot} settings. In the former, we perform experiments individually on each language by clustering the corresponding space and applying the isotropy enhancement approach.
The goal is to see whether increasing isotropy leads to performance improvement in the multilingual space and to compare the extent of improvements in cross- and multilingual tracks. 
In the zero-shot scenario, we are interested in evaluating the shared structural properties among languages, specifically, the similarity of the encoded linguistic knowledge in the dominant directions of different languages.
To this end, we obtain clusters, their means, and dominant directions on the English dataset and leverage these for isotropy enhancement in other languages.

The reported results in Table \ref{experiment-perofmance-tabel} show that increasing the isotropy in the multilingual embedding space can enhance the performance in all tracks (multi- and cross-lingual). 
The improvement could be attributed to the potential of the applied method in discarding frequency bias from the embedding space.
The visualization of the embedding space after isotropy enhancement, Figure \ref{fig:CWRs-distribution} (right), clearly reveals that the frequency bias is faded after this process.
Moreover, the results of the zero-shot setting suggest that the encoded information in dominant directions is similar across the languages because the improvement is compatible with the setting in which the dominant directions are obtained in each track individually.

\section{Conclusion}
In this paper, we provide several analyses on the geometry of multilingual embedding space from the viewpoint of isotropy. 
We show that, similarly to its monolingual English counterpart, the multilingual BERT has a highly anisotropic embedding space.
However, interestingly, the two spaces differ in their distribution of dimensions.
The English BERT has a few high-contribution outlier dimensions, whereas the multilingual space does not possess any such disruptive rouge features.
We also investigate the structure of multilingual embeddings from the perspective of frequency-based distribution and show that they have a biased structure towards word frequency and that the distribution is similar across different languages. 
As the last step, we evaluate the impact of isotropy improvement on the quality of multilingual embeddings. We observe that increasing isotropy can improve multilingual CWRs' performance on STS and address their frequency bias.

\bibliography{anthology,custom}
\bibliographystyle{acl_natbib}
\appendix
\begin{table*}
\centering
\setlength{\tabcolsep}{3.5pt}
\scalebox{0.8}{
\begin{tabular}{l c c c c c c c}
\toprule
\multicolumn{1}{c}{}            &
\multicolumn{1}{c}{\bf RoBERTa} &
\multicolumn{6}{c}{\bf XLM-R} \\

\cmidrule(lr){2-2}
\cmidrule(lr){3-8}

& {En}  & {En}  & {Es} & {Ar} & {Tr} & {Su} & {Sw}\\ 
\midrule

$I_{Cos}(\mathcal{W})$ & 0.77 & 0.96 & 0.96 & 0.96 & 0.95 & 0.95 & 0.96     \\
$I_{PC}(\mathcal{W})$  &2.5E-6 & 3.8E-11 & 3.5E-9 & 3.4E-9 & 5.3E-9 & 5.9E-9 & 5.9E-9    \\ 

\bottomrule
\end{tabular}}
\caption{\label{isotropy-table-RoBERTa} The isotropy of RoBERTa and XLM-R on a sub-set of Wikipedia, reporting based on $I_{Cos}(\mathcal{W})$ and $I_{PC}(\mathcal{W})$. }
\end{table*}
\section{XLM-R}

\label{app:XML-R}
XLM-R is a Transformer-based language model trained on 100 languages \cite{conneau-etal-2020-unsupervised}. In comparison to mBERT, it has seen much more data during its pre-training, leading to higher performance on several downstream tasks. 
For our analysis, we follow the settings used for mBERT and take RoBERTa \cite{Liu2019RoBERTaAR} as its monolingual counterpart. 

\subsection{Results}
\paragraph{Probing Isotropy.}
Table \ref{isotropy-table-RoBERTa} summarizes the isotropy evaluation for RoBERTa and XLM-R models. As we expected, both models have an extremely anisotropic distribution. However, the degree of anisotropy is much higher in the multilingual model. 
\begin{table}[t!]
\centering
\setlength{\tabcolsep}{8pt}
\scalebox{0.8}{
\begin{tabular}{lc c ccc}
\toprule
\textbf{} & \textbf{$I_{Cos}(\mathcal{W})$} && \textbf{First} & \textbf{Second} & \textbf{Third} \\ 
\midrule

RoBERTa      & 0.77       && 0.703      & 0.251      & 0.007           \\ 
\midrule

English        & 0.96       && 0.895      & 0.098      & 0.000      \\
Spanish        & 0.96       && 0.896      & 0.097      & 0.003      \\
Arabic         & 0.96       && 0.879      & 0.108      & 0.003      \\
Turkish        & 0.95       && 0.884      & 0.111      & 0.003      \\
Sundanese      & 0.95       && 0.884      & 0.097      & 0.001      \\
Swahili        & 0.96       && 0.897      & 0.088     & 0.001      \\
\bottomrule
\end{tabular}
}

\caption{\label{table-contr-RoBERTa}The contribution of top-three dimensions to the expected cosine similarity ($I_{Cos}(\mathcal{W})$) in the RoBERTa and XLM-R embedding spaces.}
\end{table}

\paragraph{Sensitivity to Rouge Dimensions.}
To find out the contribution of individual dimensions to the high cosine similarity between random embeddings and anisotropic distribution, we report the contribution of top-three dimensions in Table \ref{table-contr-RoBERTa}.

\begin{figure*}[t!]
    \centering
    \includegraphics[width=12.8cm,height=6cm]{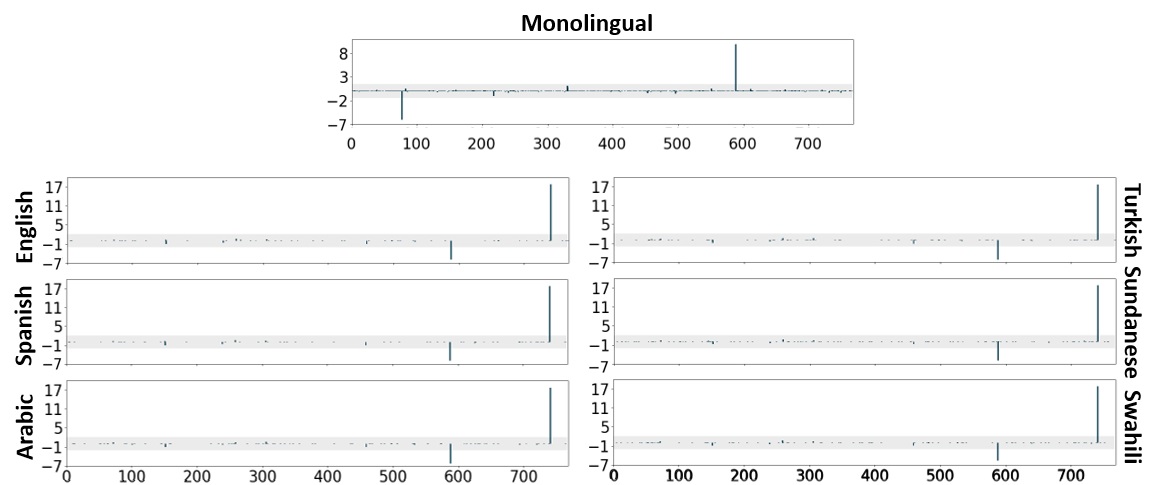}
    \caption{The average representation in RoBERTa (top) and XLM-R (bottom). The shaded area denotes $3\sigma$. Both models exhibit extreme outliers across their representations.}
    \label{fig:Outliers-RoBERTa}

\end{figure*}

In contrast to mBERT, a few dimensions have a significant role in the anisotropic distribution in the XLM-R model and RoBERTa. Furthermore, the results demonstrate that the contribution of rouge dimensions notably increases in the multilingual model.
\paragraph{Outliers.}
Following our settings in the BERT analysis (Section \ref{sec:outliers}), we plot the average representations of the RoBERTa and XLM-R models to investigate the existence of outliers. The visualization of the mean representations across different languages and the monolingual model can be found in Figure \ref{fig:Outliers-RoBERTa}. As we expected from the results of the previous part, outliers can easily be detected in the mono and multilingual models. The number of outliers is the same in both models. However, the absolute value of the outliers increases significantly in the multilingual models.
\paragraph{Conclusion.}
Our investigations on the geometry of embedding space in the RoBERTa and XLM-R models demonstrate a clear distinction to the BERT counterpart. We show that although all the mentioned models have anisotropic embedding spaces, they possess different geometrical structures from the perspective of isotropy and rouge dimensions. Working on the reasons behind such discrepancies is an interesting future direction that can enhance our knowledge of how language models shape their underlying representation space.

\section{Wordfreq}
\label{app-wordfreq}
We have employed the Wordfreq library to investigate word frequency bias in our experiments.
This library obtains word frequency from the corpus containing eight different domains in 36 languages. Our target languages are in the \emph{large} category, which means their word lists cover rare words appearing at least once per 100 million words. As a result, the wordfreq could be a suitable tool for our purpose.

\section{Frequency-based Distribution}
\label{appendix-freq}

\begin{figure}[t!]
    \centering
    \subfigure[English]{
    \includegraphics[width=3.55cm,height=2.75cm]{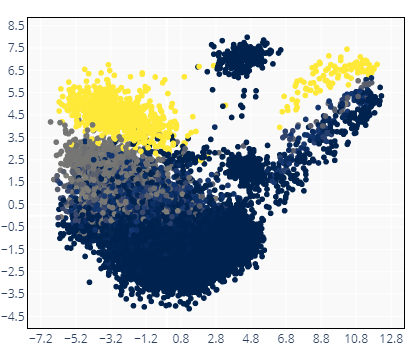}
    \includegraphics[width=3.55cm,height=2.75cm]{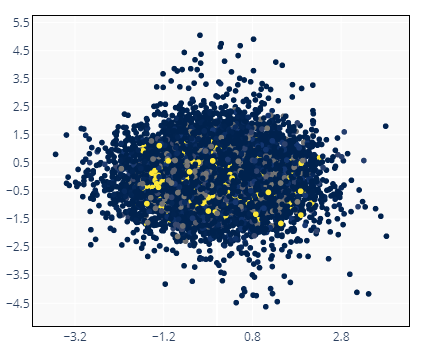}
    \label{fig:fig-geometry-en}}
    \subfigure[Arabic]{
    \includegraphics[width=3.55cm,height=2.75cm]{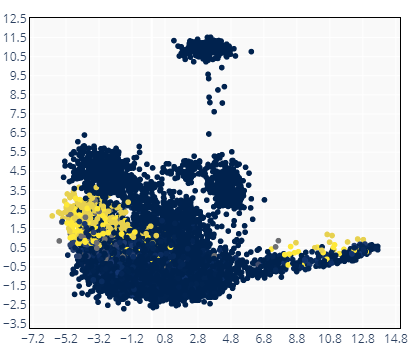}
    \includegraphics[width=3.55cm,height=2.75cm]{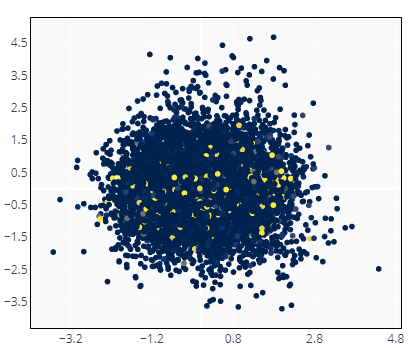}
    \label{fig:fig-geometry-es}}
    
    \caption{Degenerated (left) and isotropic (right) embedding spaces for the two languages. Frequency-based distribution can be easily detected using two top PCs in the space (lighter colors indicate higher frequency). Eliminating top dominant directions not only makes the embedding space isotropic, but also removes frequency bias in multilingual CWRs. }
    \label{fig:CWRs-distribution-app}

\end{figure}

Frequency-based distribution can negatively affect the expressiveness of space. Though it is a well-known bias in pre-trained LMs (e.g., BERT and GPT-2), it is not studied in a multilingual setting. As discussed in Section \ref{sec-frequency}, we have studied frequency bias in mBERT and demonstrated that, similarly to its monolingual counterparts, mBERT suffers from frequency-based distribution in its space. The illustration of this bias and the impact of the cluster-based approach on mitigating the issue can be found in Figure \ref{fig:CWRs-distribution-app}.

\section{Multilingual STS Task}
\label{task}
Multi and cross-lingual Semantic Textual Similarity (STS) is the main task in our experiments. STS is a paired sentence task in which samples have been labeled by a score in the continuous range of 0 (irrelevant) to 5 (most semantic similarity). In the multilingual tracks, in a pair, both sentences are in the same language, while sentences have different languages in the cross-lingual tracks. The reason behind choosing STS as the target task for our experiments is that Multilingual BERT has a pretty low performance on it.

In our experiments, we take the average of all tokens in a sentence as the sentence representation and consider the cosine similarity of the sentence representations in a sample as the semantic similarity score.

\section{Cluster-based Isotropy Enhancement}
\label{app-CIE}
We pick the cluster-based approach \citep{rajaee-pilehvar-2021-cluster} to improve the isotropy in multilingual embedding space.
In this method, the embeddings are clustered using the k-means clustering algorithm, and then dominant directions of every cluster are nulled out independently. Dominant directions have been calculated using Principal Component Analysis (PCA). The primary key in this method is obtaining dominant principal components (PCs) of clustered areas in the embedding space separately, which makes this approach suitable for exploring the clustered structure of the multilingual CWRs.
We apply the cluster-based approach to multi and cross-lingual CWRs with two different settings, Individual and Zero-shot. The number of clusters and discarded dominant directions are 7 and 12, respectively.

\end{document}